\documentclass[sigconf,natbib=true]{acmart}

\acmConference[CIKM '26]{Conference on Information and Knowledge Management}{2026}{Rome, Italy}
\acmYear{2026}
\copyrightyear{2026}

\begin{document}

\title[APEX-VW: A Document-Level EN--ES PE Dataset]{APEX-VW: A Document-Level English-Spanish Post-Editing Dataset in the Healthcare Domain}

\author{Marie Escribe}
\affiliation{%
  \institution{Universitat Polit\`ecnica de Val\`encia}
  \city{Val\`encia}
  \country{Spain}
}
\email{mcescrib@doctor.upv.es}

\author{Tharindu Ranasinghe}
\affiliation{%
  \institution{Lancaster University}
  \city{Lancaster}
  \country{United Kingdom}
}
\email{t.ranasinghe@lancaster.ac.uk}

\author{Amal Haddad Haddad}
\affiliation{%
  \institution{Universidad de Granada}
  \city{Granada}
  \country{Spain}
}
\email{amalhaddad@ugr.es}

\author{Hansi Hettiarachchi}
\affiliation{%
  \institution{Lancaster University}
  \city{Lancaster}
  \country{United Kingdom}
}
\email{h.hettiarachchi@lancaster.ac.uk}

\author{Damith Premasiri}
\affiliation{%
  \institution{Lancaster University}
  \city{Lancaster}
  \country{United Kingdom}
}
\email{d.dolamullage@lancaster.ac.uk}

\begin{abstract}
Post-Editing (PE) of Machine Translation (MT) output often involves repeating the same lexical and terminological corrections across many segments, especially in specialised and highly repetitive documents. Despite substantial work on Automatic Post-Editing (APE), most available corpora operate at the sentence level, others are synthetic, and overall not designed to study how corrections propagate in realistic Computer-Assisted Translation (CAT) workflows. This paper presents the APEX-VW (Automatic Post-Editing eXperiments on Virtual Wards) Corpus, a new open English-Spanish (EN-ES) dataset built from recent NHS virtual-ward documents and professional PE in Trados Studio, with controlled MT, terminology, and quality assurance settings. The corpus contains seven document-coherent source texts totalling 42k words, translated with four MT systems representing different paradigms and then post-edited by professional translators. Unlike prior resources such as WMT APE corpora, eSCAPE, MLQE-PE, or LangMark, the dataset preserves document order and CAT-tool context, making it suitable for research on terminology normalisation, correction propagation, and human-in-the-loop translation support. The paper describes the corpus design, data preparation, PE setup, and initial corpus statistics, and positions the resource as a benchmark for document-level APE and propagation-aware assistive tools.
\end{abstract}

\ccsdesc[500]{Information systems~Machine translation}
\ccsdesc[300]{Computing methodologies~Natural language processing}

\keywords{post-editing, automatic post-editing, post-editing dataset, machine translation, document-level machine translation, terminology, healthcare}

\maketitle

\section{Introduction}

Post-editing (PE) is now a routine part of professional translation workflows, but it often requires translators to correct repeated Machine Translation (MT) errors segment by segment ~\cite{Domingo2017segment-IMT}. This is particularly problematic in specialised domains, where a suboptimal translation of a key term may recur many times within and across documents, creating unnecessary effort and increasing the risk of inconsistency if some instances are overlooked ~\cite{Culo2017contrasting-term}. Existing Computer-Assisted Translation (CAT) tools -- such as Trados Studio or memoQ -- partly address this through Translation Memories (TMs) and Term Bases (TBs), yet they generally do not propagate newly made PE corrections beyond exact or near-exact repetition patterns.

Previous work on adaptive, interactive, and online PE has repeatedly explored the idea that systems should learn from user feedback during the translation process rather than treat each segment as an isolated unit~\cite{KnightChander1994,OrtizMartinez2010,SimardFoster2013,Lagarda2015,Chatterjee2017,Negri2018a}. From the professional perspective, user-oriented studies likewise suggest that translators value tools that can remember and reuse prior decisions, particularly in repetitive workflows where consistency is critical~\cite{Lagoudaki2008,EscribeCandelMora2024}.
At the same time, currently available PE and Automatic Post-Editing (APE) datasets do not align well with this use case. Existing resources have been crucial for the development of APE and related evaluation tasks~\cite{Bojar2015,doCarmo2020,Fomicheva2022,Velazquez2025}, but they are predominantly sentence-level and mostly designed for benchmarking rather than for studying document-level correction behaviour in realistic CAT environments. This leaves a gap for openly licensed, professionally post-edited, document-coherent datasets that make it possible to investigate correction propagation, terminology normalisation, and consistency support across a translation project.

Therefore, the present study introduces such a resource: an open English--Spanish (EN--ES) PE dataset of healthcare service documents, which are high in repetitive terminology. The corpus was designed to support research on term normalisation, repeated correction patterns, and propagation-aware assistance in realistic CAT settings. Its main contributions are: (1) a document-coherent, openly reusable EN--ES PE corpus in a high-risk healthcare domain; (2) a controlled Trados-based workflow with four MT paradigms and professional PE; and (3) detailed corpus and setup documentation to support future work on document-level APE, interactive PE, and consistency support for MT.

\section{Related Work}

APE is commonly defined as the task of improving raw MT output by learning from triplets composed of source text, raw MT output, and a corresponding post-edited version ~\cite{doCarmo2020}.
As MT, APE has evolved from rule-based approaches to statistical and neural methods~\cite{doCarmo2020}. The broader idea that MT should work in partnership with human revisers is anchored in early reflections on human--machine cooperation in translation ~\cite{BarHillel1960}. This idea was the foundations of early implementations of APE, including ~\cite{KnightChander1994} and ~\cite{AllenHogan2000}. These early approaches are particularly relevant to the present work because they already foregrounded the basic intuition behind propagation: repeated human corrections should become reusable knowledge for subsequent segments. This intuition was further developed in interactive MT research, with models designed to update incrementally from user intervention, allowing the system to adapt during the translation process rather than only offline~\cite{OrtizMartinez2010}.

More direct work on propagation emerged in PE itself. PEPr explicitly proposed propagation as a way of learning from corrections on the fly and applying them to similar future segments~\cite{SimardFoster2013}. This work paved the way for subsequent online APE systems ~\cite{Lagarda2015, Chatterjee2017, Negri2018a}. Together, these studies show that adaptive PE is most meaningful when repeated correction patterns can be captured and reused over time.

The practical importance of such assistance is also supported from the post-editor's perspective. Survey-based evidence suggests that translators value tools that can remember and reuse previous decisions~\cite{Lagoudaki2008}. More recently, a survey reported that correction propagation is among the features practitioners most want in CAT environments for PE projects~\cite{EscribeCandelMora2024}. This indicates that propagation is not merely an algorithmic convenience, but a practical requirement linked to productivity, consistency, and reduced cognitive burden in repetitive projects.

Despite this motivation, currently available corpora only partially support research in this direction. WMT APE shared-task datasets were central to the development of the field, but they were primarily designed for benchmarking and differ substantially across editions in domain, MT quality, and PE conditions~\cite{Bojar2015,doCarmo2020}.
Large synthetic resources such as eSCAPE greatly expand the amount of training data available for APE, but because they use human references as proxy post-edits, thus not capturing authentic human PE behaviour~\cite{Negri2018b}.
Newer resources have further enriched the landscape: SubEdits adds professionally post-edited subtitle data~\cite{Chollampatt2020}, MLQE-PE combines post-edits with multilingual Quality Estimation (QE) annotations~\cite{Fomicheva2022}, and LangMark expands multilingual APE benchmarking with modern MT outputs and curated triplets~\cite{Velazquez2025}.
However, these resources remain predominantly sentence-level and do not preserve contiguous document structure or CAT-workflow context, which limits their usefulness for studying terminology normalisation and correction propagation across a document.

The dataset presented in this paper is intended to complement, rather than replace, those resources. Instead of maximising scale or multilingual breadth, it prioritises document coherence, open licensing, and workflow realism through professional EN--ES PE in Trados Studio over a set of recent healthcare documents from a public entity.
This design makes it particularly suitable for studying terminology normalisation, repeated corrections, and propagation-aware assistance in conditions that are difficult to capture in shuffled sentence collections.

\section{Dataset Construction}
This section presents how the dataset was created, from the selection of source texts and the PE setup, to its availability, and also discusses ethical considerations.

\subsection{Source Selection and Preprocessing}

The dataset focuses on healthcare, specifically virtual wards and related service-delivery texts produced by the National Health Service (NHS). This domain was chosen because it combines high practical relevance with dense and repeated terminology around care pathways, hospital-at-home models, monitoring, referral, diagnostics, and organisational procedures. It also offers recent public-sector documents with clear licensing and stable provenance, making open redistribution feasible. 
After considering product documentation, instructions for use and technical documentation, the corpus design prioritised openly licensed NHS and related UK public-sector materials. This decision trades broader genre coverage for legal clarity, accessibility, and domain coherence. The final source set contains seven documents published between 2022 and 2025 and totals roughly 42k words after cleaning. Because the documents come from the same institutional ecosystem, they share conceptual framing and terminology, which is important for studying propagation not only within documents but also across the corpus as a whole. 
All source texts were converted from PDF or HTML into DOCX and cleaned conservatively for use in Trados Studio. The cleaning process preserved linguistically meaningful content and document structure while removing layout artefacts such as page furniture, navigation elements, duplicated headers, reference lists, and low-value numeric tables. This process produced source files that resemble realistic translation assignments and support stable segmentation and alignment for downstream MT and PE analysis. 

\subsection{MT and Post-Editing Setup}

The corpus was processed in Trados Studio 2024 under controlled conditions designed to approximate professional PE workflows while keeping the setup reproducible. Standard sentence-based segmentation was used, segment split/merge was allowed for local corrections, and source editing was restricted to fixing obvious artefacts inherited from source conversion. Quality assurance checks were kept close to Trados Studio defaults with minor adjustments (including terminology consistency, numeric agreement, and simple punctuation mismatches). No pre-existing TM was used at project creation, although an empty project TM was populated during work. An indicative EN-ES TB was provided to support consistency without constraining translators too rigidly. 
Four MT systems were selected to represent major paradigms currently used in localisation workflows: DeepL, ModernMT, Language Weaver, and an OpenAI-based GPT-5 system (all available as plug-ins through the RWS Appstore). All MT outputs were generated on 18 April 2026, using the then-current production models exposed through each provider's integrations in Trados Studio. Each document was associated with exactly one MT system.

Three professional linguists were involved in this phase: two carried out full PE following expectations of publishable quality, and a third focused on proofreading the resulting translations and ensuring terminological consistency. The linguists were between 25 and 40 years old, had academic training in translation and/or linguistics, and had professional experience in EN>ES translation. All had native-level competence in the target language. They were compensated according to their agreed professional rates and informed that the work formed part of a research project, but they were instructed to post-edit as they normally would in response to the translation brief. Access to the original NHS materials was provided for context and coherence checking.
The assignment was organised by document batch rather than by isolated segments. Linguist\_1 post-edited the DeepL and ModernMT batches, corresponding to Source Text (ST) ST1 and ST2, while Linguist\_2 post-edited the Language Weaver and OpenAI batches, corresponding to ST3--ST5 and ST6--ST7, respectively. This document-level allocation preserved MT consistency patterns and document flow, which is important for studying propagation-related phenomena over time.

\subsection{Availability and Ethical Considerations}

The APEX-VW dataset, including source texts, MT outputs, post-edited translations, and terminology resources, is publicly available on Zenodo at \url{https://zenodo.org/records/20457388} under a CC BY 4.0 licence (DOI: \url{https://doi.org/10.5281/zenodo.20457388}).

The design and release of APEX-VW explicitly follow the FAIR guiding principles for scientific data management and stewardship~\cite{Wilkinson2016}.

\begin{itemize}
 \item \textbf{Findable:} The dataset and accompanying terminology resources are deposited in Zenodo with a persistent DOI, rich metadata (including domain, language pair, MT systems, licences, and version), and standardised keywords.
 \item \textbf{Accessible:} All files are downloadable without registration under a clear open licence.
  \item \textbf{Interoperable:} The dataset is released in a widely used, open format (CSV) and encoded in UTF-8, with explicit field descriptions, enabling use with standard MT, APE, and CAT-tool pipelines.
  \item \textbf{Reusable:} Provenance is documented in detail (source selection, cleaning, MT configuration, PE setup), licences are stated explicitly, and a datasheet describes intended use cases, limitations, and known risks, supporting reliable reuse in future work.
\end{itemize}

Following the Datasheets for Datasets recommendations~\cite{Gebru2021}, the repository includes a datasheet that documents the motivation, composition, collection process, preprocessing, uses, and ethical considerations of APEX-VW. The datasheet details, among others, the institutional sources and licences of the NHS materials, the profiles and compensation of the post-editors, the MT and CAT configurations, and known limitations and appropriate use cases.

From an ethical and legal perspective, the dataset is based exclusively on publicly available institutional documents and professional translation work. No patient records, user-generated content, or other personal or sensitive data were collected or released.

\begin{table*}[!t]
\caption{Overview of the dataset by source text, showing corpus size, lexical profile, MT system, post-editor allocation, and MT-to-PE quality metrics.}
\label{tab:dataset-stats}
\centering
\small
\begin{tabular}{lrrrrrrrllrrr}
\toprule
ST & Words & Segs. & Avg. len. & RR & Vocab. & Lex. dens. & TTR & MT system & Post-editor & TER & BLEU & COMET \\
\midrule
ST1 & 16057 & 1110 & 14.47 & 0.21 & 2116 & 0.59 & 0.13 & DeepL           & Linguist\_1 &  9.66 & 87.09 & 0.92 \\
ST2 &  4460 &  320 & 13.94 & 0.12 & 1014 & 0.63 & 0.23 & ModernMT        & Linguist\_1 &  6.35 & 92.01 & 0.93 \\
ST3 &  5747 &  342 & 16.80 & 0.11 & 1307 & 0.62 & 0.23 & Language Weaver & Linguist\_2 & 50.14 & 35.71 & 0.80 \\
ST4 &  4029 &  247 & 16.31 & 0.14 &  828 & 0.62 & 0.21 & Language Weaver & Linguist\_2 & 52.44 & 32.29 & 0.81 \\
ST5 &  2423 &  137 & 17.69 & 0.11 &  641 & 0.64 & 0.26 & Language Weaver & Linguist\_2 & 55.47 & 30.03 & 0.79 \\
ST6 &  7320 &  424 & 17.26 & 0.13 & 1413 & 0.63 & 0.19 & OpenAI          & Linguist\_2 & 39.32 & 47.92 & 0.87 \\
ST7 &  2072 &  132 & 15.70 & 0.07 &  593 & 0.60 & 0.29 & OpenAI          & Linguist\_2 & 34.49 & 53.99 & 0.89 \\
\midrule
ALL & 42108 & 2712 & 15.53 & 0.19 & 3686 & 0.61 & 0.09 & --              & --          & 29.28 & 62.38 & 0.88 \\
\bottomrule
\end{tabular}
\end{table*}

\section{Initial Statistics and Relevance}

Table~\ref{tab:dataset-stats} summarises the main properties of the dataset at the document level, including corpus size, segment counts, lexical profile, MT system assignment, post-editor allocation, and PE outcome metrics. In total, the corpus contains 42,108 words and 2,712 analysed segments across seven document-coherent source texts, with an average segment length of 15.53 words, an overall Repetition Rate (RR) of 0.19, a vocabulary size of 3,686, a lexical density of 0.61, and a type--token ratio (TTR) ranging from 0.13 to 0.29\footnote{The corpus-level TTR value (0.09) is lower than any per-document TTR because TTR decreases as text length increases, therefore the corpus-level value is not directly comparable to the individual document values.}. These figures confirm that the dataset combines moderate lexical variety with substantial repetition, making it suitable for studying terminology normalisation and propagation across segments and documents. 

Inspired by the use of RR proposed by ~\cite{Bertoldi2013RR} and later used as a complexity indicator in the WMT APE shared tasks~\cite{Bhattacharyya2023WMTAPE}, we define RR for APEX-VW as a measure of the repetitiveness of a text based on the rate of non-singleton \(n\)-gram types for \(n = 1 \dots 4\) and combining them using the geometric mean.
Formally, for each \(n \in \{1,2,3,4\}\), let \(V_n^{>1}\) be the set of \(n\)-gram types that occur more than once in the text and \(V_n\) the set of all \(n\)-gram types.
The RR is then defined as
\[
\mathrm{RR} = \Bigg( \prod_{n=1}^{4} \frac{\lvert V_n^{>1} \rvert}{\lvert V_n \rvert} \Bigg)^{\!1/4} .
\]

A clear variation appears across documents. RR values indeed range from 0.07 to 0.21, with the highest value observed in ST1 and the lowest in ST7, while average segment length ranges from 13.94 to 17.69 words. This variation is useful because it creates different propagation conditions, from shorter, more repetitive texts to longer and lexically denser segments that may require more context-sensitive corrections. 

The PE metrics suggest substantial differences in the amount of intervention required for different MT outputs, although these values should not be interpreted as a controlled ranking because MT systems were assigned by document rather than evaluated on the same source text.
At the corpus level, PE yielded TER 29.28~\cite{Snover2006}, BLEU 62.38~\cite{Papineni2002}, and COMET 0.88~\cite{Rei2020}. DeepL and ModernMT show lower TER and higher BLEU/COMET on their respective texts, whereas Language Weaver and OpenAI required more extensive editing on the documents assigned to them. For the intended use of the dataset, the important point is not system comparison as such, but the fact that the corpus captures a range of PE effort profiles and therefore a range of propagation-relevant correction patterns.

Another contribution of the resource is the expansion of the terminology material during PE. The initial indicative TB was deliberately lightweight (21 terms, 29 acronyms), but the PE process produced an updated TB with 215 term entries and a separate acronym resource with 79 entries, which were consolidated and validated during the final consistency-checking step by the proofreading linguist. These expanded resources are valuable in their own right because they record terminology decisions that emerged during the workflow and can support future research on terminology management, consistency modelling, and terminology-aware MT and APE. 
In addition, a further analysis of edit statistics shows that PE involved substantial rewriting rather than only minimal surface correction. Across the corpus, 6,790 insertions, 1,285 deletions, and 9,523 replacements were implemented, with a mean edit distance of 6.49 per segment. The number of PE tokens (57,659) is also higher than the number of MT tokens (52,154), which suggests that many edits involved explicitation, restructuring, or expansion rather than simple substitution. This is relevant for propagation research because it shows that repeated interventions are not limited to single-word replacements but may involve richer correction patterns. 
Finally, the corpus supports at least two propagation scenarios identified during dataset preparation. First, in some cases, MT is internally consistent but repeatedly wrong, so the same incorrect translation must be corrected multiple times across a document (Scenario~A). Second, in other cases, MT itself is inconsistent, and the post-editor must normalise competing translations of the same concept across the document set (Scenario~B). To illustrate Scenario~A, we inspected how the English term \emph{virtual ward} was translated within ST2. In this document, \emph{virtual ward} (and its plural form) occurs 35 times in the source across 31 segments, and has one main rendering in Spanish, namely \emph{sala virtual} (or \emph{salas virtuales}). In other words, MT is lexically consistent but systematically chooses a suboptimal term (as the linguists chose to use \emph{unidad de hospitalización virtual}, so corrections had to be implemented repeatedly wherever the term appears. Scenario~B is exemplified by ST1, where \emph{virtual ward(s)} occurs frequently (254 source instances) but the MT system distributes these occurrences across several competing Spanish translations. The most common options include \emph{sala virtual} (219), \emph{unidad virtual} (15) and \emph{servicio de hospitalización virtual} (6), alongside a longer tail of rarer variants such as \emph{distrito virtual}, \emph{sistema virtual} and \emph{unidad de atención a distancia}. In this setting, the post-editor is not only correcting individual instances, but also normalising terminology by converging these alternatives onto a preferred form throughout the document.

\section{Limitations}

It should be acknowledged that the dataset is restricted to one language direction, one domain cluster, a modest number of documents, and a small number of post-editors. It also reflects a snapshot of MT systems and PE decisions at a specific moment in time rather than a longitudinal record of evolving edits. In addition, because different source documents were assigned to different MT systems, the current corpus is better suited to resource release and methodological study than to strict cross-system benchmarking. Moreover, the CAT setup targets a specific tool (Trados Studio) and configuration, which may limit direct transfer of observations to other environments.
These limits are balanced by the dataset’s main strength: it provides an openly reusable, professionally post-edited, document-level EN-ES resource specifically designed for correction propagation research in realistic CAT conditions. By combining document coherence, open licensing, MT diversity, and detailed workflow documentation, it fills a gap left by sentence-level datasets and creates a basis for work on propagation-aware PE assistance, document-level MT and APE, and consistency support in specialised translation.

\section{Future Work}

Several avenues for future work are opened by this dataset. First, the corpus can support the design and evaluation of propagation-aware assistance in CAT tools, for example by learning when and how to propose document-level term propagation or pattern-based suggestions during PE. Second, the resource enables systematic experiments on document-level and online APE, including models that exploit repeated correction patterns and terminology edits to adapt over the course of a project. Third, CAT environments could incorporate internal adaptive models that learn not only from term-level corrections but also from recurring stylistic and register-related edits. Fourth, extending the dataset to additional domains, language pairs, and CAT environments would make it possible to test how robust propagation strategies are across different workflows. Finally, combining the existing corpus with richer annotation---such as error typologies, fine-grained effort measures, or explicit propagation events---would support more detailed studies of how human post-editors manage consistency over time.

\section{Conclusion}

This study has presented APEX-VW, an openly available EN--ES PE dataset in the healthcare domain, constructed from recent NHS virtual-ward and related service documents.
By preserving document structure, CAT-workflow context, and detailed MT and PE statistics, the corpus fills a gap left by predominantly sentence-level APE resources.
The dataset is specifically designed to support research on correction propagation, terminology normalisation, and consistency support in realistic professional settings, and is intended to serve as a basis for new propagation-aware models and tools in document-level translation workflows.

\section*{Acknowledgements}

This project is funded by the European Association for Machine Translation (EAMT) through its Sponsorship of Activities programme. We would also like to express our sincere gratitude to Paloma Vega Centeno for her precious help during the post-editing phase.

\section*{GenAI Usage Disclosure}

The present paper and the accompanying dataset documentation were prepared with the assistance of Generative AI tools. These tools were used to help with language polishing, restructuring of existing text, and formatting suggestions, based on content and instructions provided by the authors. All writing suggestions were reviewed and approved by the authors. All dataset design decisions, experimental configurations and analyses are original work by the authors.


\bibliographystyle{ACM-Reference-Format}
\bibliography{refs} 

\end{document}